\documentclass[11pt,letterpaper]{article}

\usepackage[left=1in,right=1in,top=1in,bottom=1in]{geometry}
\usepackage[utf8]{inputenc}
\usepackage[T1]{fontenc}

\usepackage{amsmath,amssymb,amsthm,mathtools}
\usepackage{bm}
\usepackage{mathrsfs}
\usepackage{mathabx}
\usepackage{extarrows}
\usepackage{upgreek}

\usepackage[dvipsnames,table]{xcolor}
\usepackage{graphicx}
\usepackage{subfigure} 
\usepackage[figurename=Figure,font=footnotesize,labelfont=bf]{caption}
\usepackage{xltabular}
\usepackage{tikz}
\usepackage[edges]{forest}
\usetikzlibrary{arrows.meta,positioning,fit,calc}

\usepackage{booktabs}
\usepackage{multirow}
\usepackage{longtable}
\usepackage{threeparttable}
\usepackage{array}
\usepackage{tabularx}
\usepackage{makecell}

\usepackage{type1cm}
\usepackage{lettrine}
\usepackage{moreverb}
\usepackage{framed}
\usepackage{wrapfig}
\usepackage{paralist}
\usepackage{indentfirst}
\usepackage{relsize}
\usepackage[normalem]{ulem}
\usepackage{mwe}

\usepackage{algorithmic}
\usepackage{tikz}
\usepackage{forest}
\usetikzlibrary{arrows.meta,positioning,fit,calc}

\usepackage[dvipsnames,table]{xcolor}

\usepackage[sort&compress,numbers]{natbib}
\usepackage{authblk}

\usepackage[
  bookmarks=true,
  bookmarksopen=true,
  bookmarksnumbered=true,
  colorlinks=true,
  linkcolor=blue,
  citecolor=blue,
  filecolor=blue,
  urlcolor=blue
]{hyperref}

\hypersetup{
  unicode=false,
  pdftoolbar=true,
  pdfmenubar=true,
  pdffitwindow=false,
  pdfstartview={FitH},
  pdftitle={My title},
  pdfauthor={Author},
  pdfsubject={Subject},
  pdfcreator={Creator},
  pdfproducer={Producer},
  pdfkeywords={keywords},
  pdfnewwindow=true
}

\definecolor{impl}{RGB}{121,85,72}
\definecolor{orch}{RGB}{0,150,136}
\definecolor{solv}{RGB}{63,81,181}
\definecolor{auto}{RGB}{255,152,0}
\definecolor{disc}{RGB}{0,121,107}
\definecolor{catone}{RGB}{33,150,243}
\definecolor{cattwo}{RGB}{76,175,80}

\graphicspath{{./Figure/}}

\providecommand{\keywords}[1]{\textbf{\textit{Keywords: }} #1}

\begin{document}

\title{\textbf{Large language models for\\ partial differential equation workflows}}

\author[1]{Han Wan}{}
\author[1]{Rui Zhang}{}
\author[1]{Hao Sun}{}

\affil[1]{\small Gaoling School of Artificial Intelligence, Renmin University of China, Beijing 100872, China}

\date{}

\maketitle

\normalsize

\vspace{-18pt} 
\begin{abstract}
\small

Numerical methods and deep learning have advanced the modelling and simulation of systems governed by partial differential equations (PDEs), while formulating equations, configuring solvers, and translating simulations into design and control decisions continue to demand substantial domain expertise and computational resources. Large language models (LLMs) offer opportunities to automate and coordinate these tasks by combining scientific knowledge, mathematical reasoning, code generation, and tool use. Here we review LLM-assisted PDE research across three stages, including discovery, solving, and optimization. We examine how LLMs interact with physical knowledge, numerical methods, and computational feedback to formulate models, implement solution procedures, and guide design and control. We also review existing benchmarks and evaluation practices for assessing scientific validity, computational performance, and workflow reliability. Finally, we discuss three interconnected challenges and opportunities: developing mathematical and physical reasoning tailored to specific PDE problems; building transferable domain expertise through domain-specific and multimodal post-training; and integrating active experimental design with scientific validation for open-ended problems. These directions outline a path towards a more substantive role for LLMs in developing and testing scientific ideas and computational methods across PDE research.

\end{abstract}


\vspace{12pt} 

\section{Introduction}

In the eighteenth century, d’Alembert described the motion of a vibrating string using the wave equation~\cite{https://doi.org/10.1111/1600-0498.12171}, helping to establish the theory of partial differential equations (PDEs). This pioneering work demonstrated how concise mathematical expressions can capture complex physical phenomena. After that, Maxwell's equations~\cite{fitzpatrick2008maxwell} unified electricity, magnetism, and light, laying the theoretical foundations for wireless communication and radar; and the Navier--Stokes equations~\cite{temam2001navier} enabled quantitative descriptions of fluid motion that now guide the design of aircraft, turbines, and propulsion systems. Ranging from microscopic to macroscopic scales, PDEs have become central to scientific advances that transform our understanding of nature and underpin the technologies of everyday life~\cite{Evans2010PDE,colizza2007reaction,liu2023experimental,de2024pattern}.

\begin{figure}[!t]
    \centering
    \includegraphics[width=1\linewidth]{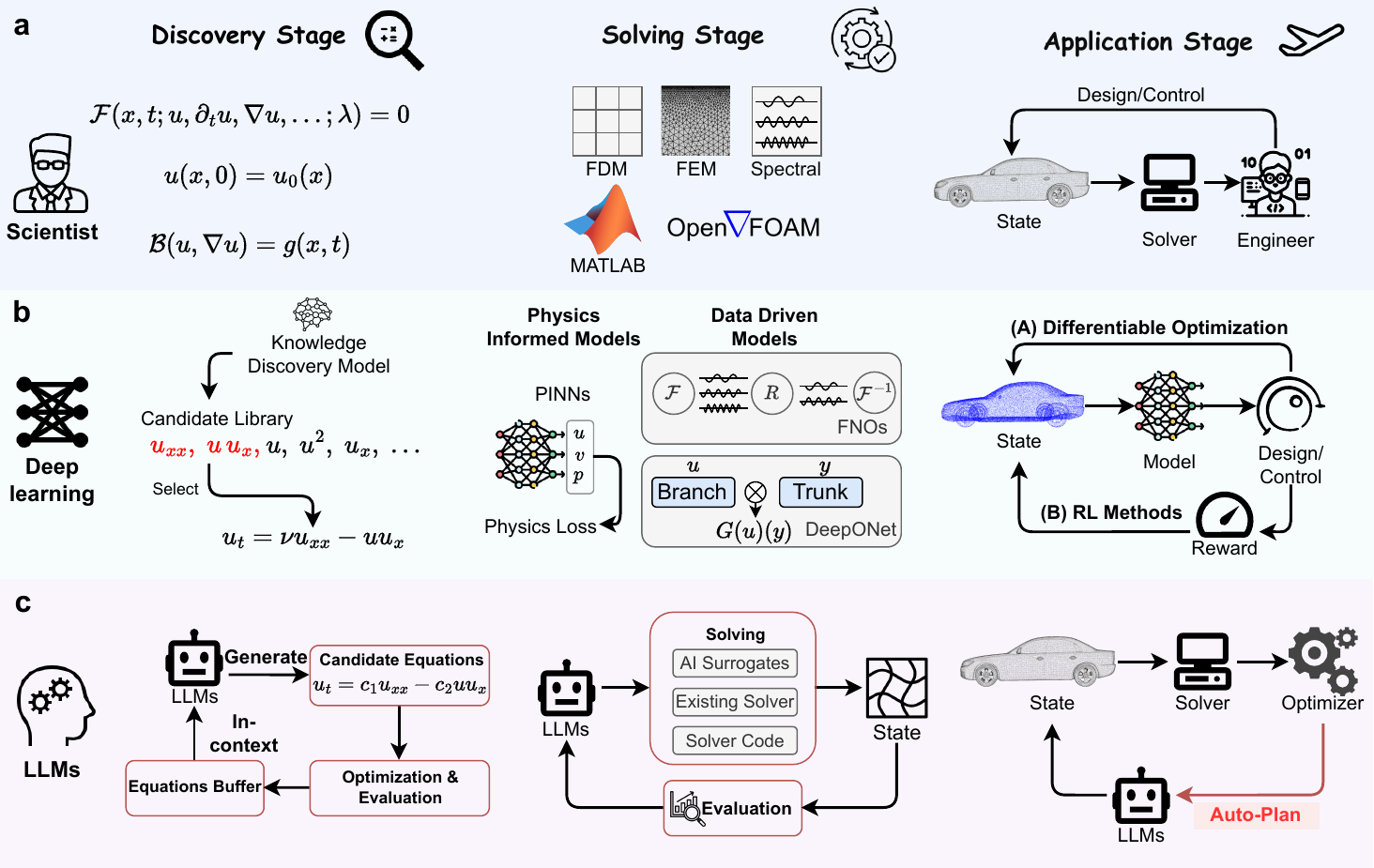}
\caption{\textbf{Comparison of traditional, deep-learning, and LLM-assisted workflows for PDE scientific computing.} The workflow is organized into three stages: \textbf{Discovery}, \textbf{Solving}, and \textbf{Optimization}. \textbf{(a)} Traditional PDE scientific computing establishes the methodological foundation of the workflow but relies heavily on expert knowledge and manual coordination across stages. \textbf{(b)} Deep-learning approaches enhance individual components, including equation discovery, neural PDE solvers, surrogate modelling, and differentiable optimization, while remaining largely confined to specific tasks or workflow stages. \textbf{(c)} LLM-assisted systems coordinate heterogeneous scientific knowledge, executable tools, numerical solvers, and iterative feedback across the complete workflow, providing workflow-level assistance from model discovery and executable computation to downstream scientific and engineering applications.}
    \label{fig:pde_workflow_bridge}
\end{figure}

Applying PDEs to real-world problems requires substantial expertise in translating physical assumptions into mathematical models, selecting and implementing numerical methods, and judging whether the results are physically meaningful (Fig.~\ref{fig:pde_workflow_bridge}\textbf{a}). Governing equations are typically established by mathematicians, physicists, and domain scientists from first principles through detailed and rigorous reasoning, and are then tested and refined against empirical observation. Because analytical solutions are generally unavailable for practical PDE problems, various numerical methods have been developed to approximate their solutions, including finite-difference, finite-volume, finite-element, and spectral methods~\cite{LeVeque2007FiniteDifference,versteeg2007introduction,BrennerScott2008FEM,Trefethen2000Spectral}. In practical simulations, these methods require expert choices in meshing, time stepping, and discretization; large-scale problems further impose substantial computational costs, requiring efficient implementation and parallel computation~\cite{Antonietti2016DG,Vivarelli2025Fluids}. Beyond forward simulation, PDE solutions can guide changes to a system's controls or design to achieve a desired objective, with applications including flow control, inverse design, and topology optimization~\cite{Lions1971OptimalControl,Hinze2009PDEConstraints,Gunzburger2002FlowControl,BendsoeSigmund2004TopologyOptimization}. In practice, achieving these objectives often requires repeated PDE solves together with expert configuration and adjustment, resulting in high computational and operational costs in complex settings~\cite{Oberkampf2002VV}.

Over the past decade, deep-learning methods have used the expressive power of neural networks to automate or accelerate many computational components of traditional PDE workflows (Fig.~\ref{fig:pde_workflow_bridge}\textbf{b}). Data-driven discovery methods, such as sparse regression and related learning approaches, recover governing equations or candidate terms from measurements~\cite{brunton2016discovering,rudy2017data,chen2021physics}. For forward problems, neural networks can accelerate or complement traditional solvers by learning solution representations from data and physical constraints. Physics-informed neural networks provide mesh-free formulations for complex geometries and high-dimensional PDE problems by embedding governing equations and boundary conditions into training objectives~\cite{han2018solving,raissi2020hidden,lu2021physics,yu2022gradient}. Neural operators learn mappings between function spaces, enabling fast prediction across families of PDE problems and extending learned solution operators to irregular geometries and unstructured discretizations~\cite{lu2021learning,li2021fourier,tran2023factorized,wan2026sino}. For optimization and control, deep-learning-based methods can accelerate forward analysis through surrogate models, while also supporting PDE-constrained decision-making by learning or differentiating through control objectives~\cite{Rabault2019DRLFlowControl,Holl2020DifferentiablePhysics, sun2026gano}. Despite these advances, using deep learning to accelerate PDE workflows still requires substantial involvement from researchers with expertise in both scientific computing and artificial intelligence, including selecting suitable model architectures, incorporating physical constraints, tuning optimization procedures, and assessing numerical accuracy.

In recent years, large language models (LLMs), exemplified by ChatGPT~\cite{openai2024gpt4technicalreport}, DeepSeek~\cite{deepseekai2025deepseekv3technicalreport}, and Gemini~\cite{geminiteam2025geminifamilyhighlycapable}, have expanded the capabilities of AI from language generation to multimodal understanding, supporting applications across scientific research and engineering. These advances have motivated researchers to explore how LLMs can support different stages of PDE research, with the aim of reducing manual intervention and improving computational efficiency (Fig.~\ref{fig:pde_workflow_bridge}\textbf{c}). 
By combining learned scientific knowledge with capabilities in code generation~\cite{chen2021codex}, structured reasoning~\cite{wei2022chain}, and tool use~\cite{karpas2022mrkl}, LLMs can connect natural-language specifications, mathematical representations, computational software, and execution feedback. In the \textbf{Discovery Stage}, LLMs draw on learned knowledge to propose candidate equations, compare physical hypotheses, and formulate mathematical models from problem descriptions. Coupled with data fitting and physical checks, these capabilities can guide the search towards plausible model structures and support the iterative revision of hypotheses~\cite{du2024llm4ed,feng2025physpde}. In the \textbf{Solving Stage}, their knowledge of programming patterns and numerical methods supports solver-code generation and software configuration, while their ability to interpret execution feedback enables iterative debugging and revision. These capabilities can reduce the manual effort required to implement, test, and refine numerical workflows for specified PDE problems~\cite{li2026codepde,yue2026foamagent,Fazliani2025PDESHARP}. In the \textbf{Optimization Stage}, LLMs interpret control and design requirements to formulate objectives and constraints, propose candidate solutions, and coordinate their evaluation using numerical solvers. By incorporating simulation feedback into subsequent proposals, they can help automate the repeated formulation, evaluation, and refinement steps involved in PDE-constrained search~\cite{soroco2025pdecontroller,zhang2025using}.

Recent studies have explored the use of LLMs across different stages of PDE research, creating a need to bring these advances together and clarify what they offer to scientific computing. In this Review, we organize existing work into three stages: discovery, solving, and optimization. For each stage, we examine how LLMs combine scientific knowledge, reasoning, and tool use with numerical methods, and clarify the tasks they perform within the broader workflow. We then review existing benchmarks and evaluation practices, considering how they assess mathematical and physical validity, computational performance, and workflow reliability. Finally, we discuss three interconnected challenges and opportunities: reasoning that adapts to individual PDE problems, post-training that develops transferable domain expertise, and experimental design and scientific validation for open-ended questions.

\section{A Workflow Taxonomy of LLM Frameworks for PDEs}
\label{sec:taxonomy}

Consider a physical field $u(x,t)$ defined over a spatial domain $\Omega$ and a time interval $[0,T]$, whose governing equation can be written as
\begin{equation}
\mathcal{F}\left(x,t;u,\partial_t u,\nabla u,\ldots;\lambda\right)=0,
\label{eq:general_pde}
\end{equation}
where $\mathcal{F}$ denotes the governing operator and $\lambda$ denotes system or control parameters. The problem is completed by initial conditions $u(x,0)=u_0(x)$ and boundary conditions $\mathcal{B}(u,\nabla u)=g(x,t)$ on $\partial\Omega$. Building on this representation, the \textbf{Discovery--Solving--Optimization} workflow is organized according to the mathematical task addressed at each stage and the role played by the LLM.

The \textbf{Discovery} stage concerns identifying or formulating the mathematical structure of the PDE problem. Traditionally, such structure is often derived by scientists from first principles, such as conservation laws and symmetries. In this stage, LLMs can support the identification of the operator $\mathcal{F}$ from observations and physical priors. Furthermore, once the structural form of the governing PDE is specified, LLMs can assist with PDE-level mathematical reasoning, such as deriving analytical representations, identifying solution structures, or analyzing qualitative and asymptotic behavior. We therefore distinguish two main Discovery-stage directions: \emph{LLM-assisted equation discovery} and \emph{LLM-assisted problem formulation and mathematical reasoning}.


The \textbf{Solving} stage aims to compute the solution $u(x,t)$ when the PDE formulation (Eq.~\eqref{eq:general_pde}) is given. LLMs can support this process by generating solver code, configuring and orchestrating numerical software, assisting deep-learning-based PDE solvers, or interpreting diagnostic feedback such as the PDE residual, convergence behavior, stability violations, and runtime failures. We distinguish three main Solving-stage directions, including \emph{LLM-augmented numerical software workflows}, \emph{LLM-based PDE solver code generation}, and \emph{LLM-assisted deep-learning PDE solvers}.

The \textbf{Optimization} stage concerns PDE-centred workflows that seek the best control, design, or model choices under a specified objective and the governing PDE. Given an objective functional $\mathcal{L}$, the goal can be written abstractly as
\begin{equation}
\min_{\lambda} \mathcal{L}\bigl(u,\lambda\bigr)
\quad
\text{s.t.}
\quad
\mathcal{F}\left(x,t;u,\partial_t u,\nabla u,\ldots;\lambda\right)=0,
\label{eq:pde_optimization_overview}
\end{equation}
where $\lambda$ represents physical parameters, control variables, or other design variables. The objective functional $\mathcal{L}$ quantifies the performance of a candidate control, design, or model choice. LLMs assist this stage by formulating objectives and constraints, proposing or refining candidate parameters or designs, coordinating solver-in-the-loop evaluations, and interpreting optimization or data-fitting feedback.

The three stages therefore correspond to different unresolved components of the same mathematical workflow: the \textbf{Discovery} stage identifies or formulates $\mathcal{F}$ and related problem specifications, the \textbf{Solving} stage computes the state $u(x,t)$, and the \textbf{Optimization} stage optimizes $\lambda$ with respect to a task-level objective $\mathcal{L}$ under the governing PDE constraint. Table~\ref{tab:llm4pde-taxonomy} provides an article-level mapping of representative studies onto this workflow taxonomy, further organizing them by methodological direction, article type, and the principal function performed by the LLM.

\begingroup
\footnotesize
\setlength{\tabcolsep}{3.0pt}
\renewcommand{\arraystretch}{1.14}
\rowcolors{2}{gray!5}{white}

\begin{longtable}[!t]{@{}
>{\raggedright\arraybackslash}p{0.23\textwidth}
>{\raggedright\arraybackslash}p{0.18\textwidth}
>{\centering\arraybackslash}p{0.07\textwidth}
>{\raggedright\arraybackslash}p{0.49\textwidth}@{}}

\caption{Representative LLM-based systems and language- or symbol-conditioned studies in PDE workflows.}
\label{tab:llm4pde-taxonomy}\\

\toprule
Study & Taxonomy category & Year & Main focus \\
\midrule
\endfirsthead

\toprule
Study & Taxonomy category & Year & Main focus \\
\midrule
\endhead

\bottomrule
\endfoot

\bottomrule
\endlastfoot

\multicolumn{4}{@{}l}{\textit{LLMs for the Discovery stage}} \\
\midrule

KeplerAgent~\cite{yang2026thinklikescientistphysicsguided}
& Equation discovery
& 2026
& After a poor initial fit, calls an algorithmic symmetry detector, converts the detected symmetry into an exact constraint, and reruns the equation-fitting tool. \\

LaSR~\cite{grayeli2024lasr}
& Equation discovery
& 2024
& Learns reusable structural guidance by contrasting better and worse equations, then uses it to create, modify, and combine candidates during symbolic regression. \\

LLM4PD~\cite{luo-etal-2025-large-language}
& Equation discovery
& 2025
& Uses language models to select and review equation terms, then proposes coefficients for numerical simulation; a stalled coefficient search returns to term selection. \\

LLM-SR~\cite{shojaee2025llmsr}
& Equation discovery
& 2025
& Proposes executable equation structures, while an external numerical optimizer fits their parameter placeholders to data and the resulting scores guide later prompts. \\

PiSR~\cite{taskin2026knowledgeintegration}
& Equation discovery
& 2026
& Scores candidate expressions for dimensional consistency, simplicity and physical plausibility; these scores enter an external symbolic-regression objective alongside data-fit and complexity terms. \\

LLM-EPDE~\cite{ivanchik2025llmepde}
& Equation discovery
& 2025
& Generates candidate PDE expressions as Python functions from observed fields and their derivatives; external coefficient fitting, residual evaluation and complexity assessment guide subsequent proposals. \\

IGSR~\cite{saveliev2026igsr}
& Equation discovery
& 2026
& Proposes nonlinear equation terms, while numerical regression fits their weights and term-removal scores guide pruning; the retained and discarded terms inform later proposals within a tree search. \\


LLM4ED~\cite{du2024llm4ed}
& Equation discovery
& 2024
& Alternates score-guided equation revision with LLM-generated combinations and mutations; parsing, coefficient fitting, scoring, and candidate retention remain external. \\

DrSR~\cite{wang2025drsr}
& Equation discovery
& 2025
& Describes patterns in the observations and summarizes lessons from successful, unsuccessful, or invalid equation attempts; both forms of memory guide later proposals. \\

SR-Scientist~\cite{xia2026srscientist}
& Equation discovery
& 2026
& Uses the language model as the search controller: it chooses analyses and equation tests, then revises hypotheses from numerical feedback across many tool-use steps. \\

EqGPT~\cite{xu2025generativediscoverypartialdifferential}
& Equation discovery
& 2025
& Trains a GPT-style decoder on handbook equations, generates free-form PDE structures, and iteratively adapts it using candidates scored by external data fitting. \\

PiT-PO~\cite{wang2026llmbasedscientificequationdiscovery}
& Equation discovery
& 2026
& Updates the equation generator during search, rewarding data fit and physical consistency while penalizing unnecessary or redundant terms. \\

Bhatnagar et al.~\cite{bhatnagar2025equations}
& Scientific reasoning
& 2025
& Predicts which mathematical operators a PDE solution may require, thereby narrowing the operator set searched by a downstream symbolic solver without constructing the solution itself. \\

Agentic Symbolic Search~\cite{yu2026agenticsymbolicsearchcharacterizing}
& Scientific reasoning
& 2026
& Searches readable solution representations for a known PDE; external gradient optimization fits each proposed program and a fixed evaluator scores its equation and condition errors. \\

\midrule
\multicolumn{4}{@{}l}{\textit{LLMs for the Solving stage}} \\
\midrule

OpenFOAMGPT 1.0~\cite{Pandey_2025}
& Numerical software workflows
& 2025
& Builds executable OpenFOAM cases from natural-language requests, grounds them in retrieved tutorials, and revises failed configurations using solver error logs. \\

MetaOpenFOAM 1.0~\cite{chen2024metaopenfoamllmbasedmultiagentframework}
& Numerical software workflows
& 2024
& Coordinates planning, tutorial-grounded file generation, run-script writing, external OpenFOAM execution, and error-directed file replacement from a natural-language CFD request. \\

AutoCFD (NL2FOAM dataset)~\cite{Dong2025}
& Numerical software workflows
& 2025
& Fine-tunes a 7B model on synthetic natural-language-to-OpenFOAM examples, then combines it with input checking, solver execution, and error-driven language-model revision. \\

Foam-Agent~\cite{yue2026foamagent}
& Numerical software workflows
& 2026
& Automates OpenFOAM setup, meshing, execution, repair, and visualization using stage-specific tutorial retrieval, dependency-aware file generation, and lessons from prior failed attempts. \\

ChatCFD~\cite{fan2026chatcfd}
& Numerical software workflows
& 2026
& Converts article or dialogue requirements and an external mesh into an OpenFOAM case, then uses execution errors, retrieved files, and failed-repair lessons to coordinate corrections. \\

MechAgents~\cite{ni2024mechagents}
& Numerical software workflows
& 2024
& Divides planning, mechanics formulation, FEniCS coding, and critique among GPT-4 agents, using solver feedback and peer review to repair elasticity simulations. \\

AutoFEA~\cite{hou2025autofea}
& Numerical software workflows
& 2025
& Retrieves planning and code examples from successful finite-element projects, generates a CalculiX input file step by step, reviews it against the request, and then runs it. \\

MCP-SIM~\cite{park2026mcpsim}
& Numerical software workflows
& 2026
& Maintains a shared record of specifications, code versions, execution traces, and fixes while iteratively building and explaining simulations across several physics domains. \\

CFDAgent~\cite{xu2025cfdagent}
& Numerical software workflows
& 2025
& Uses language models to route and configure geometry preparation, flow simulation, and requested analyses; external generation, meshing, solver, and plotting tools produce the scientific artifacts. \\

ALL-FEM~\cite{DEOTALE2026118985}
& Numerical software workflows
& 2026
& Combines models fine-tuned on executable, expert-checked FEniCS scripts with coordinated planning, formulation, coding, critique, and execution-feedback repair. \\

OpInf-LLM~\cite{wang2026opinllm}
& Numerical software workflows
& 2026
& Parses PDE requests and either calls fixed reduced-model tools or generates compact solver code; numerical routines learn the reduced dynamics and reconstruct solutions at new parameters. \\

AI CFD Scientist~\cite{somasekharan2026aicfdscientistopenended}
& Numerical software workflows
& 2026
& Runs end-to-end CFD studies by modifying case-specific OpenFOAM code, compiling and simulating it, checking rendered fields and reference metrics, and drafting a cited paper. \\

CodePDE~\cite{li2026codepde}
& Solver code generation
& 2026
& Generates Python PDE solvers at inference time, repairs crashes or invalid numbers, refines promising candidates from measured performance, and selects the best sampled program. \\

AutoNumerics~\cite{du2026autonumericsautonomouspdeagnosticmultiagent}
& Solver code generation
& 2026
& Builds inspectable solver programs through staged parsing, method screening, code generation, coarse-to-fine execution and repair, and metric-based selection; external runs supply errors and residuals. \\

PDE-SHARP~\cite{Fazliani2025PDESHARP}
& Solver code generation
& 2025
& Uses staged PDE analysis to guide solver generation, then lets three model judges shortlist candidates without execution and iteratively patch nominees using optional numerical feedback. \\

AutoPDE~\cite{dong2026autopde}
& Solver code generation
& 2026
& Generates PDE solver code while keeping formulation, discretization, stabilization, linear solver, preconditioner, and resolution choices explicit and revisable from pilot-run evidence. \\

PDE-Agent~\cite{liu2025pdeagent}
& Solver code generation
& 2025
& Orchestrates existing PDE tools from natural-language requests; language models plan and validate calls, while external tools compute, train, visualize, and share intermediate objects. \\

LLM-PDEveloper~\cite{wu2025automatedcodedevelopmentpde}
& Solver code generation
& 2025
& Extends an existing Python lattice-Boltzmann library with reusable solver or boundary-condition modules instead of generating a one-off simulation script. \\

PINNsAgent~\cite{wuwu2025pinnsagent}
& Deep-learning PDE solvers
& 2025
& Automates architecture and hyperparameter search for physics-informed neural networks, seeding trials from similar equations and updating the search with externally measured error. \\

Lang-PINN~\cite{he2026langpinn}
& Deep-learning PDE solvers
& 2025
& Converts informal physics requests into modular neural PDE solvers by combining equation consensus, history-guided architecture selection, symbolic loss checks, training diagnostics, and localized repair. \\

ADSL-PDE~\cite{kong2026adslpde}
& Deep-learning PDE solvers
& 2026
& Constrains language-model search to a solver-design format that can be checked and compiled deterministically into trainable neural PDE solvers. \\

ICON-LM~\cite{yang2025iconlm}
& Deep-learning PDE solvers
& 2025
& Uses a decoder-only numerical operator model to infer a new function mapping from demonstrations and optional text, then evaluate it without retraining for that mapping. \\

PITT~\cite{lorsung2024pitt}
& Deep-learning PDE solvers
& 2024
& Conditions a neural PDE predictor---not a general-purpose LLM---on tokenized equations, using learned attention updates to correct the underlying operator's field prediction. \\

PROSE-PDE~\cite{sun2025prosepde}
& Deep-learning PDE solvers
& 2025
& Uses a purpose-built numerical--symbolic Transformer, not a general-purpose LLM, to forecast future states and recover governing equations across several PDE families. \\

PDEformer-1~\cite{ye2024pdeformer1}
& Deep-learning PDE solvers
& 2024
& Uses a graph- and symbol-conditioned neural solver, not an LLM, to represent a complete one-dimensional PDE and predict its continuous solution field. \\

\midrule
\multicolumn{4}{@{}l}{\textit{LLMs for the Optimization stage}} \\
\midrule

PDE-Controller~\cite{soroco2025pdecontroller}
& Optimization
& 2025
& Formalizes PDE-control goals, proposes intermediate goals, and generates programs whose finite-element optimization and constraint scoring run outside the language models. \\

OptMetaOpenFOAM~\cite{chen2026optmetaopenfoam}
& Optimization
& 2026
& Uses a language model to plan and explain CFD studies while OpenFOAM, sampling, response fitting, sensitivity analysis, plotting, and external optimizers perform the numerical work. \\

Zhang et al.~\cite{zhang2025using}
& Optimization
& 2025
& Uses in-context language-model inference to choose each new population's center, while external flow and heat-transfer solvers score the sampled shapes. \\

Sun et al.~\cite{sun2026selfevolvingscientificagentdiscovers}
& Optimization
& 2026
& Evolves readable feedback-control code for a two-joint swimmer using earlier simulated trajectories and flows; an external fluid solver supplies behavior and scores. \\

AgenticSciML~\cite{jiang2026agenticsciml}
& Solver-design optimization
& 2026
& Evolves scientific machine-learning architectures and training strategies through specialized-agent debate, method retrieval, executable evaluation, error repair, and score-guided search. \\

COSMO-Agent~\cite{deng2026cosmoagent}
& Engineering optimization
& 2026
& Reinforcement-trains an 8B language model to turn finite-element feedback into parametric design revisions until constraints are met or the interaction budget expires. \\

TurboAgent~\cite{du2026turboagent}
& Engineering optimization
& 2026
& Coordinates compressor design by turning requirements and evaluated results into workflow and search decisions; learned geometry models, predictors, optimizers, and CFD or structural tools perform the computations. \\

TO-Master~\cite{lin2026tomaster}
& Engineering optimization
& 2026
& Maps conversational design intent to checked topology-optimization inputs; external geometry, meshing, finite-element, sensitivity, and numerical-optimization tools generate and optimize the design. \\

\end{longtable}

\endgroup

\section{LLMs for the Discovery Stage}
\label{sec:discovery_stage}

\begin{figure}[!t]
    \centering
    \includegraphics[width=0.99\linewidth]{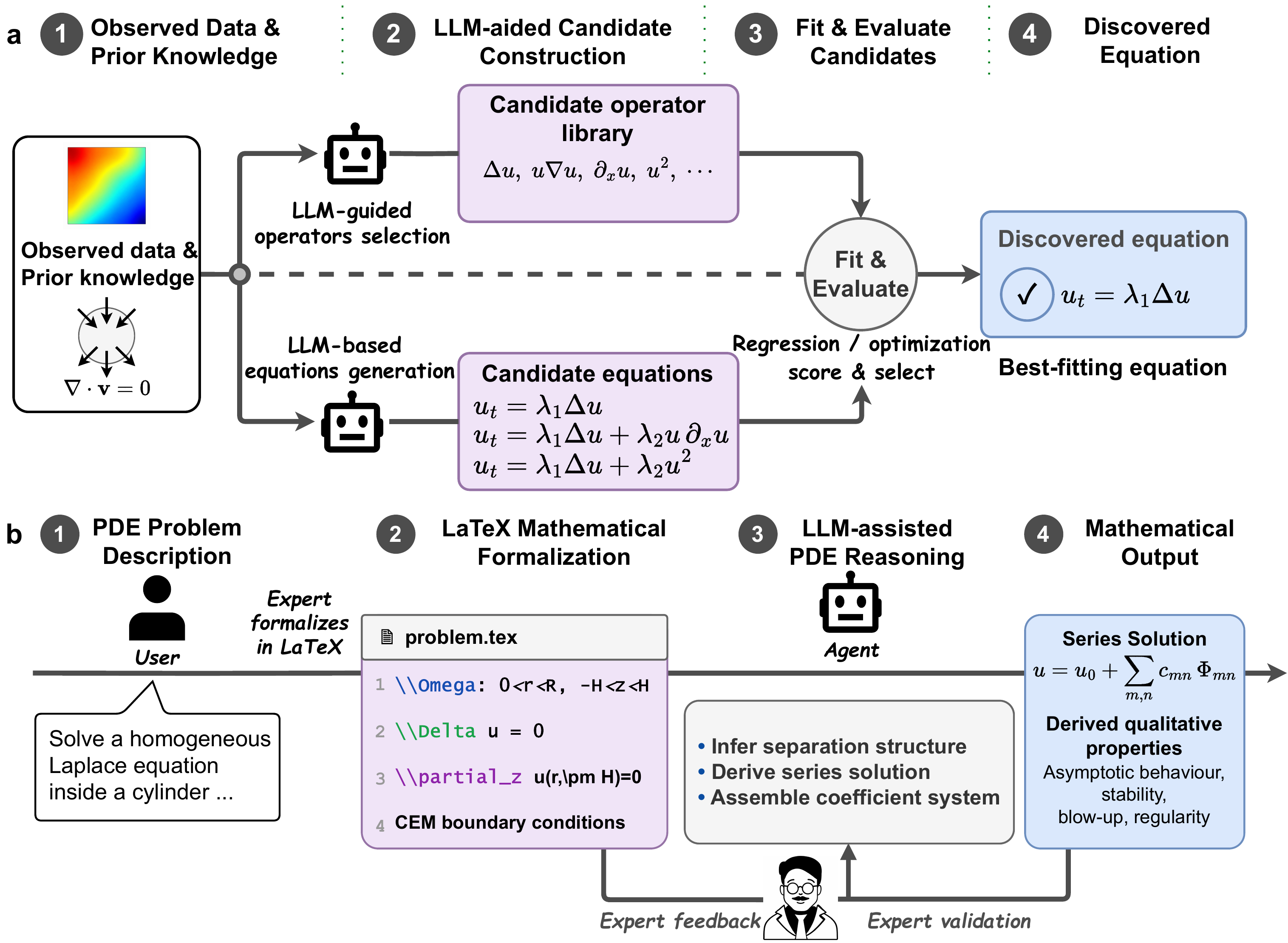}
    \caption{\textbf{Representative LLM-assisted workflows in the Discovery stage.}
    \textbf{a}, LLM-assisted equation discovery. Observations and prior scientific knowledge guide either the construction of a candidate operator library or the direct generation of candidate equations. Candidate structures are then fitted, evaluated, and selected through regression, optimization, simulation, or other validation procedures before a reasonable governing equation is retained.
    \textbf{b}, LLM-assisted problem formulation and scientific reasoning. A PDE problem is first formalized as a mathematical specification, after which an LLM supports structured PDE reasoning, derivation of solution representations, and qualitative analysis under expert feedback and validation.}
    \label{fig:fig2}
\end{figure}

The Discovery stage encompasses two main directions. Equation discovery starts from observational data and seeks to infer the governing mathematical relation. PDE problem formulation and reasoning start from a scientific question or a specified mathematical structure and seek to construct a precise mathematical description or characterize the problem's analytical and qualitative properties. LLMs support both directions by connecting scientific knowledge with symbolic representations and mathematical reasoning, as illustrated by the representative workflows in Fig.~\ref{fig:fig2}.

\subsection{LLM-assisted equation discovery}

Equation discovery aims to identify the governing PDE of a physical system from observational data. A common approach is to construct a library of candidate operators and use a regression or search engine to identify the terms and coefficients that best explain the observations. Deep-learning methods can accelerate parts of this search and fitting process, while candidate libraries and admissible mathematical structures remain largely expert-designed. As illustrated in Fig.~\ref{fig:fig2}\textbf{a}, LLMs extend this process by incorporating scientific and physical knowledge into candidate-space construction, structural selection, and direct equation generation.

For the first path, LLMs guide an external equation-search process by shaping the candidate space and refining candidate equations with scientific and numerical feedback. A key step in conventional equation discovery is determining which mathematical terms or structures should be considered during the search. LLMs can incorporate physical knowledge into this process to guide the selection of candidate structures, thereby directly influencing the efficiency and reliability of subsequent regression or symbolic-search procedures. KeplerAgent~\cite{yang2026thinklikescientistphysicsguided} makes this process explicit by first using an LLM together with physics-based tools to infer symmetries from observational data. These inferred symmetries are then translated into function-library choices and structural constraints that directly restrict the equation space explored by symbolic-regression engines such as PySINDy~\cite{desilva2020pysindy} and PySR~\cite{cranmer2023pysr}. LaSR~\cite{grayeli2024lasr} similarly uses LLM-derived information to guide an external symbolic search, but obtains this guidance from the search history itself: the LLM abstracts reusable natural-language concepts from high-performing symbolic hypotheses, and these concepts are used to steer subsequent evolutionary search. As new high-quality hypotheses emerge, they are further distilled into higher-level concepts, allowing knowledge accumulated during the search to continuously influence later candidate generation. KeplerAgent and LaSR both use LLM-derived information to guide an external symbolic search. LLM4PD~\cite{luo-etal-2025-large-language} extends this involvement beyond search guidance to a coarse-to-fine refinement process. In the coarse phase, an LLM selects candidate PDE terms from a predefined library and an LLM-based review agent checks the proposed structure. In the fine phase, the selected structure is retained while the LLM iteratively refines its coefficients using the coefficient--error trajectory returned by a PDE solver.  PiSR~\cite{taskin2026knowledgeintegration} uses an LLM to score expressions generated by symbolic-regression engines for dimensional consistency, simplicity and physical plausibility. These scores contribute an auxiliary penalty to the search objective, allowing domain knowledge to influence candidate selection alongside numerical fit and structural complexity. These approaches use LLM-derived scientific knowledge, search experience, and numerical feedback to guide or refine an external equation-search process.

For the second path, language models themselves participate directly in the generation and iterative refinement of candidate equation structures. In LLM-SR~\cite{shojaee2025llmsr}, high-performing equation programs stored in a dynamic experience buffer are reused as in-context examples to guide subsequent generations. LLM4ED~\cite{du2024llm4ed} also exploits elite historical candidates, but uses them to directly modify existing equations or generate new structures through evolutionary operations such as recombination and structural mutation. LLM-EPDE~\cite{ivanchik2025llmepde} supplies the LLM with compact representations of observed fields and their numerically computed derivatives, together with previously evaluated candidates. The model generates candidate PDE expressions as Python functions, while external coefficient fitting, equation-residual evaluation and complexity assessment guide subsequent proposals. Other methods make the feedback more diagnostic by identifying why current candidates succeed or fail. DrSR~\cite{wang2025drsr} combines structural analysis of the observed data with reflection over successful, unsuccessful, and invalid generations, allowing both data characteristics and previous search outcomes to shape subsequent proposals. RESTART~\cite{li2026restart} analyzes residual patterns left unexplained by the current equation and converts them into targeted structural corrections, while storing repeatedly useful refinements in a reusable structure library. IGSR~\cite{saveliev2026igsr} makes this feedback specific to individual equation terms. An LLM proposes nonlinear terms, numerical regression fits their weights, and changes in validation error upon term removal guide pruning. Records of retained and discarded terms then inform further proposals within a Monte Carlo tree search. SR-Scientist~\cite{xia2026srscientist} goes beyond the common generate--evaluate--refine loop by allowing the model to actively determine what analysis should be performed before revising a candidate equation. It can write and execute diagnostic code to inspect the data and test current hypotheses, and then use the resulting evidence to decide how the equation should be modified.

EqGPT~\cite{xu2025generativediscoverypartialdifferential} takes a more explicitly generative approach by pretraining on PDE structures collected from mathematical handbooks and learning statistical patterns among mathematical operators and terms. Generated PDE candidates are evaluated and fed back into a generation--evaluation--optimization loop to further improve subsequent generation. Importantly, EqGPT was applied to real experimental measurements of strongly nonlinear surface gravity waves approaching breaking, where it proposed a previously unreported governing PDE, extending LLM-based equation generation from benchmark recovery to real-world scientific discovery. PiT-PO~\cite{wang2026llmbasedscientificequationdiscovery} further updates the equation generator itself through reinforcement learning, using physics-validity constraints and token-level regularization to steer the policy toward valid and parsimonious equation structures.

These studies connect scientific knowledge with numerical evidence to guide candidate construction, evaluation and revision, and in some cases the acquisition of additional observations. Complementing the discovery of governing equations from data, LLMs can also help formulate scientific problems and reason about the mathematical structure and properties of specified PDEs.

\subsection{LLM-assisted problem formulation and scientific reasoning}

Beyond equation discovery, LLMs can help formulate scientific problems and reason about the physical and mathematical structure of PDEs. As illustrated in Fig.~\ref{fig:fig2}\textbf{b}, scientific questions, modelling settings and specified PDEs provide starting points for organizing assumptions and constraints, selecting mathematical representations and identifying tractable subproblems. These activities support the development of analytical descriptions and arguments about solution behaviour.

At the model-formulation level, LLMs can assist in constructing and revising physical models within an existing PDE framework. Yang et al.~\cite{yang2025largelanguagemodeldriven} use an LLM to develop near-wall turbulence models for flows affected by adverse pressure gradients, rotation and surface roughness. The LLM proposes model formulations based on relevant physical mechanisms, while wall-stress and velocity predictions together with CFD simulations provide feedback for subsequent revision. This creates an iterative modelling process in which physical reasoning is translated into mathematical model components and refined against numerical evidence.

Once a PDE problem has been specified, LLMs can assist mathematical analysis by reasoning through derivations and proposing structures for solution expressions. In Garnadi's case study~\cite{garnadi2025llm}, the LLM receives the Laplace equation, cylindrical geometry and boundary conditions, and helps develop intermediate derivation steps and expressions for the electrical potential. Bhatnagar et al.~\cite{bhatnagar2025equations} use an LLM to predict the mathematical operators likely to appear in a solution expression from the equation, forcing and boundary conditions. These operators then constrain the search for an analytical approximation. Agentic Symbolic Search~\cite{yu2026agenticsymbolicsearchcharacterizing} uses PDE theory, problem constraints, and previous search results to guide an agent in generating and revising candidate solution representations. These representations are implemented as differentiable symbolic programs, allowing their continuous parameters to be fitted by gradient-based optimization. A fixed evaluator assesses PDE residuals, initial and boundary conditions, and initial-time compatibility to guide subsequent revisions. For a radially symmetric Keller--Segel problem describing chemotactic aggregation, this process yields a nine-parameter approximation that captures density concentration in a shrinking core over the evaluated time window.

LLMs also contribute to theorem-level reasoning by suggesting intermediate lemmas, counterexamples and proof strategies. Liu et al.~\cite{liu2026homogenization} report using a ChatGPT-supplied lemma in a convergence proof for periodic homogenization of viscous quadratic Hamilton--Jacobi equations, relating a model with fine-scale periodic structure to an effective macroscopic description. LLM-suggested constructions have also supported counterexamples to gradient bounds based only on ellipticity~\cite{le2026unbounded} and to convergence as viscosity tends to zero in a mechanical Hamilton--Jacobi ergodic problem~\cite{liu2026nonexistence}. At the level of overall proof strategy, reported contributions include characterizing energy-minimizing shapes in Gamow's liquid-drop model~\cite{chodosh2026liquiddrop}. For a flow that drives sustained magnetic-field growth, Rowan~\cite{rowan2026dynamo} reports that an LLM-generated proof idea reduces the key analysis to a recursion involving three selected Fourier modes. In these studies, the authors report developing and checking the model-generated contributions as part of the final mathematical proofs.

A larger-scale example is OpenAI's reported result for the Navier--Stokes existence and smoothness problem, a foundational question in mathematical fluid dynamics and one of the Clay Mathematics Institute's Millennium Prize Problems~\cite{claynavierstokes}. OpenAI reports finite-time blowup for the three-dimensional incompressible Navier--Stokes equations with smooth forcing: the flow starts from rest, and its velocity becomes unbounded while its kinetic energy remains bounded~\cite{openai2026navierstokes}. A coordinated multi-agent system explored alternative proof and counterexample directions, with approximately 10,000 concurrent agents in the group that produced the result. Intermediate mathematical results were consolidated across groups and incorporated into follow-up prompts to guide further exploration. The resulting argument was subsequently formalized and verified in Lean~\cite{openai2026navierstokesannouncement}.

These studies demonstrate a substantial expansion in the mathematical research capabilities of LLMs, from assisting with intermediate derivations and arguments through expert interaction to generating original mathematical constructions, key proof strategies and complex proofs. The Navier--Stokes result reported by OpenAI is particularly significant: through sustained parallel exploration and the integration of intermediate results, a multi-agent system produced a solution to a fundamental problem that had challenged the mathematical community for nearly a century. This achievement shows that LLM systems can make original breakthroughs at the frontier of mathematical research. Applying these mathematical reasoning and multi-agent collaboration capabilities more broadly to scientific discovery, with model-generated physical hypotheses and candidate laws tested against experimental data and iteratively refined through feedback, has the potential to accelerate the discovery and understanding of natural laws.

\section{LLMs for the Solving Stage}
\label{sec:stage_solving}

\begin{figure}[!t]
    \centering
    \includegraphics[width=0.79\linewidth]{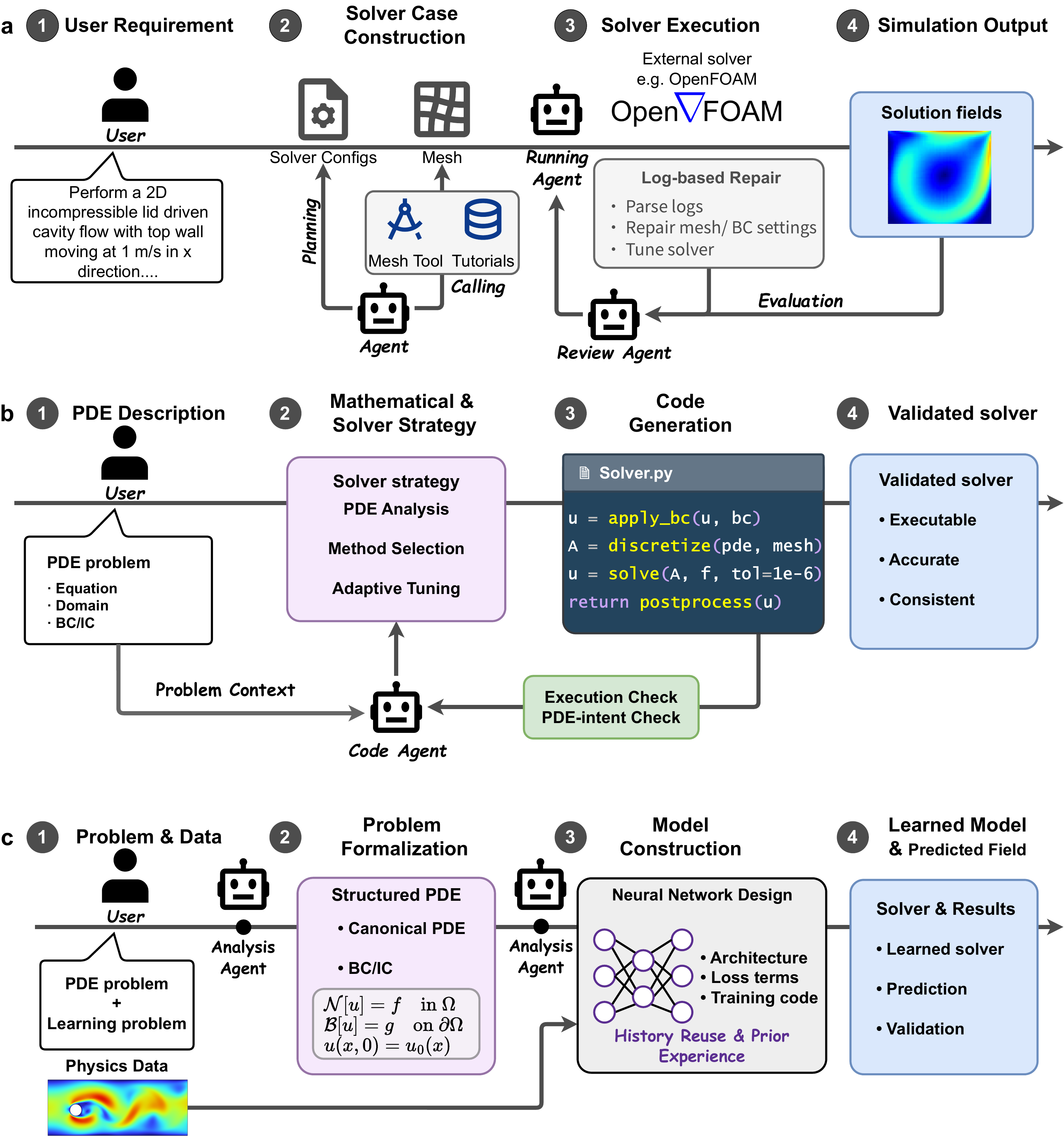}
    \caption{\textbf{LLMs for PDE solving-stage workflows.}
    \textbf{a}, In numerical software workflows, LLM agents translate PDE specifications into solver configurations, execute external tools, and revise failed cases using logs and diagnostics.
    \textbf{b}, In solver-code generation, LLMs generate, test, and debug numerical programmes from mathematical or natural-language PDE descriptions.
    \textbf{c}, In deep-learning PDE solvers, LLMs assist architecture design, loss construction, training workflows, or language-conditioned PDE representations.}
    \label{fig:solving_stage}
\end{figure}

Once the governing equations and conditions of a PDE problem have been specified, its mathematical formulation must be translated into a reliable computational process to support scientific research and engineering applications. Conventional numerical methods typically require experts to select discretization schemes, implement solvers or configure existing software, and specify meshes, time steps and convergence criteria. These tasks demand substantial expertise and considerable manual effort in configuration and debugging. Deep learning surrogate models can accelerate repeated simulations after training, but their predictive performance depends on the coverage of the training data, often requiring further adaptation and validation for unseen boundary conditions, geometries or parameter regimes. Drawing on numerical methods, software documentation and computational feedback, LLMs can apply mathematical reasoning, code generation and tool use to support solver construction, configuration and debugging. They can also assist in designing, training and validating learned solvers, including incorporating equation and boundary-condition information into these models. These capabilities can lower technical barriers and improve the efficiency and adaptability of PDE-solving workflows.

As summarized in Fig.~\ref{fig:solving_stage}, we distinguish three complementary directions according to the primary role of the LLM: configuring and orchestrating established numerical software, generating numerical solver implementations from PDE specifications, and assisting the construction and conditioning of learned PDE models. These directions are discussed as \emph{LLM-augmented numerical software workflows}, \emph{LLM-based PDE solver code generation}, and \emph{LLM-assisted deep-learning PDE solvers}, respectively.

\subsection{LLM-augmented numerical software workflows}
\label{sec:solving_traditional_solver_workflows}

A direct way for LLMs to support PDE solving is to invoke and coordinate established numerical simulation software. As illustrated in Fig.~\ref{fig:solving_stage}\textbf{a}, LLMs combine knowledge of physics and numerical methods with tool use to translate mathematical and physical requirements into coordinated choices for meshing, numerical schemes, solver settings and post-processing. They then run the solver and use execution feedback to diagnose problems and revise configurations, ultimately obtaining the numerical solution.

Representative OpenFOAM-based systems study how LLMs can invoke established CFD software from natural-language simulation requirements. Early systems combine retrieval from existing OpenFOAM~\cite{jasak2007openfoam} cases or tutorials with task decomposition, configuration generation, simulation execution, and iterative repair, enabling LLMs to translate high-level CFD requirements into executable solver cases~\cite{Pandey_2025,chen2024metaopenfoamllmbasedmultiagentframework}. Moreover, AutoCFD~\cite{Dong2025} fine-tunes Qwen2.5-7B-Instruct on NL2FOAM, a dataset of 28,716 natural-language--OpenFOAM configuration pairs, and then embeds the adapted model in an end-to-end OpenFOAM workflow. Subsequent systems extend this paradigm from case generation to more complete and reliable CFD workflows. MetaOpenFOAM 2.0 and OpenFOAMGPT 2.0~\cite{Chen2025MetaOpenFOAM2.0,feng2025openfoamgpt20endtoendtrustworthy} broaden the workflow to multi-stage CFD tasks, including simulation, analysis and visualization. A further line focuses on reliability and failure recovery: Foam-Agent~\cite{yue2026foamagent} retrieves workflow-specific reference cases from structured tutorial indices to keep coupled settings such as turbulence models and numerical schemes consistent, while ChatCFD~\cite{fan2026chatcfd} localizes runtime-error files, identifies dependent configurations that must be modified together, and uses accumulated repair histories to correct persistent failures. OpenFOAM primarily uses the finite volume method, whereas software based on other numerical methods, such as the finite element method implemented in FEniCS~\cite{alnaes2015fenics}, may be better suited to particular PDE problems and modelling requirements.

LLM-assisted automation has also been explored on these alternative platforms, supporting both the construction of simulation code and the configuration and execution of numerical solvers. MechAgents~\cite{ni2024mechagents} uses collaborating agents to formulate elasticity problems, express the governing equations in weak form, and generate, execute and revise FEniCS simulation code. To strengthen such code-generation capabilities, ALL-FEM~\cite{DEOTALE2026118985} expands expert-written FEniCS examples into more than 1,000 verified scripts for domain-specific fine-tuning and incorporates the adapted models into a multi-agent workflow for finite-element code generation. Simulation setup also requires geometry, material properties and other problem conditions to be represented as usable inputs. CFDagent~\cite{xu2025cfdagent} converts textual or visual inputs into three-dimensional geometries and coordinates meshing, immersed-boundary flow simulation and post-processing. PDE-Agents~\cite{adhikari2026pdeagentsllmorchestratedmultiagentframework} draws on a GraphRAG knowledge base containing material properties, known failure patterns and simulation histories to guide agents in configuring and invoking a FEniCSx~\cite{baratta2023dolfinx} backend. Across successive execution and revision cycles, MCP-SIM~\cite{park2026mcpsim} uses persistent shared memory to record problem specifications, execution feedback and modification histories, enabling agents to refine simulations using accumulated context. Together, these studies illustrate how LLMs translate problem descriptions, domain knowledge and execution feedback into simulation code and configurations, and coordinate iterative collaboration to carry out and refine numerical simulations.

The same orchestration approach can be applied to reduced-order numerical models. OpInf-LLM~\cite{wang2026opinllm} uses an LLM to extract the equation, parameter values and initial and boundary conditions from natural-language requests and translate them into tool calls. These calls invoke polynomial regression to estimate reduced operators for the requested parameters from operators fitted offline, followed by numerical integration of the reduced-order dynamical system. The resulting reduced trajectories are mapped back to PDE solution fields using a basis previously constructed from solution data.

Beyond automating simulations, AI CFD Scientist~\cite{somasekharan2026aicfdscientistopenended} uses numerical-software automation for more open-ended CFD investigation, connecting literature-grounded task formulation, model modification, simulation execution, and vision-language inspection. In a periodic-hill study, the system identified a runtime correction to the Spalart--Allmaras model that reduced the lower-wall $C_f$ RMSE against DNS by 7.89\%. Its vision-language verification module also detected 14 of 16 planted failures that were not identified by solver-level checks. This example demonstrates the potential of LLMs to coordinate CFD software workflows for automated fluid-mechanics research, moving beyond prescribed simulation execution toward model revision and physical-credibility assessment.
\subsection{LLM-based PDE solver code generation}

Although established numerical software supports a wide range of simulations, its existing algorithms and interfaces can limit flexibility for particular PDE problems. LLM-based solver code generation provides a complementary approach by constructing numerical implementations tailored to the mathematical structure and computational demands of a problem. As illustrated in Fig.~\ref{fig:solving_stage}\textbf{b}, this workflow begins with a specification of the governing equations, computational domain, and initial and boundary conditions. The LLM analyzes the problem, selects a numerical strategy, and implements the solution procedure as executable code. Evaluations of execution errors, numerical accuracy, stability, residuals, and computational performance then guide revisions to both the implementation and the numerical strategy. This process connects mathematical problem descriptions with solver implementations that can be tested and refined through computation.

Developing such workflows requires both the ability to generate numerical programs and mechanisms for assessing the numerical choices behind them. Evaluations by Kashefi and Mukerji~\cite{kashefi2023chatgpt} demonstrate that general-purpose LLMs can generate and modify numerical programs for problems including Poisson and Navier--Stokes equations, while also documenting failures such as singular matrices and incomplete code generation. These results highlight the value of coupling code generation with explicit numerical design and computational checks. AutoNumerics~\cite{du2026autonumericsautonomouspdeagnosticmultiagent} translates a problem description into a structured specification and proposes candidate schemes for spatial discretization and, where needed, time integration. Agents shortlist these schemes, generate their implementations, and test them from coarse to finer resolutions, using solution-error or PDE-residual evaluations to inform the final selection. AutoPDE~\cite{dong2026autopde} similarly treats numerical design as a process that can be revised through feedback. It maintains an explicit record of discretization, stabilization, and solver settings, and uses error and runtime measurements from low-cost pilot runs to calibrate these choices and revise their implementation. Together, these approaches make the numerical decisions arising from an LLM's interpretation of a PDE available for examination and adjustment.

Once a numerical strategy has been implemented, computational feedback can guide code repair and further improvements in accuracy. CodePDE~\cite{li2026codepde} combines reasoning-guided code generation with iterative debugging and numerical refinement. It translates a PDE specification into a complete solver implementation, uses runtime error traces to guide code repair, and evaluates errors against reference solutions to support further improvement. Generating and evaluating multiple candidates also allows the framework to select more accurate implementations. This search introduces an additional concern: repeated solver executions can consume a substantial evaluation budget. PDE-SHARP~\cite{Fazliani2025PDESHARP} addresses this concern by using LLM-based assessment to select promising candidates for execution. It also proposes code patches that combine ideas from alternative implementations, directing numerical evaluation towards selected candidates and using the resulting feedback to guide subsequent refinement.

Mathematical derivations and existing example code provide another route to constructing specialized solvers. The LLM-empowered CAE agent~\cite{guo2025largelanguagemodelempowerednextgeneration} develops reduced-order solvers for parametric PDEs using Tensor-decomposition-based A Priori Surrogates (TAPS). Given a target PDE and detailed derivations and code for a simpler example, the LLM derives the weak formulation and matrix equations, then adapts the template into a solver implementation for the new problem. Verification tasks are incorporated into this workflow: the agent generates symbolic code for manufactured solutions and corresponding forcing terms, and invokes tools to conduct convergence studies and evaluate numerical errors. The resulting TAPS solvers construct solution approximations directly from the governing equations without requiring offline training snapshots. This example illustrates how mathematical reasoning, template adaptation, and computational testing can be combined within a specific numerical framework.

The scope of code generation can also extend from individual solver implementations to reusable numerical software. LLM-PDEveloper~\cite{wu2025automatedcodedevelopmentpde} takes mathematical and algorithmic descriptions as input and modifies the source code of XLB, a lattice Boltzmann library. Its agents develop solver components and accompanying test programs, then inspect, execute, and revise the implementation. Tasks include implementing new solvers, adding reaction terms to advection--diffusion equations, and introducing boundary-condition modules. Here, the generated code extends the capabilities of an existing library, making new numerical functionality available within a reusable software framework.

Across both standalone solvers and library components, mathematical verification can complement execution-based testing by examining how an implementation relates to its intended PDE. Oschatz et al.~\cite{oschatz2025verification} investigate a partly automated verification chain connecting floating-point code to its discrete numerical algorithm and then to the continuous PDE. For a pseudospectral solver of the one-dimensional periodic Poisson equation, they combine program analysis with mathematical derivations to bound discretization and floating-point errors. This work adds an analytical perspective to solver assessment, linking properties of the implemented computation to the mathematical problem it approximates.

LLM-based solver code generation offers a highly flexible approach to numerical PDE solving. By combining existing mathematical tools and numerical methods, LLMs can construct and optimize problem-specific solvers while accommodating requirements such as GPU deployment and differentiable computation. Through iterative search and validation, these systems can outperform expert-written baseline implementations in accuracy or computational efficiency on specific tasks. These capabilities can reduce repetitive work in implementation, debugging and platform adaptation, enabling experts to devote more attention to numerical methods and scientific research while opening new opportunities for automated solver development and algorithmic innovation.

\subsection{LLM-assisted deep-learning PDE solvers}
\label{sec:solving_deep_learning}

To harness deep-learning PDE solvers for accelerated repeated simulations, LLMs are employed to automate their development and setup. As illustrated in Fig.~\ref{fig:solving_stage}\textbf{c}, the workflow starts from a PDE problem together with available physical or observational data. The LLM formalizes the problem into a structured mathematical specification and translates it into solver-design choices, such as network architecture, loss formulation, sampling strategy, and optimization procedure. Prior setups and training experience further inform these decisions, while execution and validation results provide feedback for subsequent revision.

As a widely adopted deep-learning approach, Physics-informed neural networks (PINNs)~\cite{raissi2019physics} approximate PDE solutions by embedding governing equations into the loss function. Applied to PINNs, LLM frameworks help navigate their sensitivity to architecture, loss weighting, and numerical precision through iterative code generation and validation. PINNsAgent~\cite{wuwu2025pinnsagent} uses physics-guided knowledge replay to retrieve effective configurations from previously solved PDEs with similar mathematical and physical characteristics, and uses memory tree reasoning to guide subsequent architecture and hyperparameter search. Candidate configurations are implemented and trained, with errors measured against reference solutions informing further exploration. Under the reported five-iteration search setting, it achieves the lowest or joint-lowest reported mean MSE on 12 of 14 PDEs compared with random and Bayesian search. Lang-PINN~\cite{he2026langpinn} constructs symbolic PDEs from natural-language descriptions, selects architectures using PDE characteristics and prior results, and generates modular PINN training code. Execution errors and training diagnostics guide targeted revisions of the affected modules. PDE-Agent~\cite{liu2025pdeagent} organizes PINN construction and training through a modular DeepXDE toolchain, using graph memory and a shared resource pool to maintain tool dependencies and intermediate computational objects during execution and error correction.

Beyond PINNs, LLM assistance has also been extended to broader neural PDE solver workflows and reduced-order models. MyCrunchGPT~\cite{Kumar2023} integrates core neural solvers (e.g., PINNs and DeepONet~\cite{lu2021learning}) with auxiliary modules for geometry processing and solution verification, extending natural-language interaction beyond a single PINN model. ADSL-PDE~\cite{kong2026adslpde} further applies LLMs to the automatic design of neural PDE solvers: it lets the LLM adjust key design choices within a structured design space, and then compiles the resulting design into executable code to reduce implementation failures caused by unrestricted code generation.

More open-ended efforts begin to involve LLM agents in the development of the solution method itself. AgenticSciML~\cite{jiang2026agenticsciml} uses multiple LLM agents to retrieve methodological knowledge, propose and critique neural-network architectures, loss functions, and training procedures, and implement candidate methods as executable code. Numerical experiments then evaluate the resulting solvers and provide feedback for subsequent revision. In its Burgers example, the system develops a training procedure that first fits the initial and boundary conditions and subsequently introduces the governing-equation residual. Rather than only configuring a predefined solver, this workflow allows the agent to explore alternative model and training designs through iterative computational feedback.

Furthermore, learned PDE solvers incorporate textual or multimodal problem descriptions~\cite{zhouunisolver,negrini2025multimodalpdefoundationmodel}, with Text2PDE~\cite{zhou2025text2pde} and ICON-LM~\cite{yang2025iconlm} illustrating text-conditioned solution generation and in-context operator learning, respectively. Symbolic approaches~\cite{lorsung2024pitt,sun2025prosepde}, including PDEformer-1~\cite{ye2024pdeformer1}, encode governing equations as token sequences or computational graphs to guide solution prediction. A related line of work~\cite{zhu2024fluidllmlearningcomputationalfluid,Yang2023}, exemplified by UPS~\cite{shen2024ups}, adapts language-model architectures or pretrained representations to PDE and fluid dynamics prediction.

Existing studies demonstrate that LLMs can support numerical PDE solving by invoking, configuring and tuning deep-learning models. Current efforts largely focus on existing models and predefined design spaces, while autonomous exploration of new deep-learning solution methods remains at an early stage. As LLM systems develop greater capabilities for self-improvement, future agents may draw on the mathematical structure of PDEs and computational feedback to accumulate solution experience and autonomously propose, implement and validate new model architectures and training methods, advancing the development of deep-learning PDE solvers.

\section{LLMs for the Optimization Stage}
\label{sec:optimization_stage}

\begin{figure}[!t]
    \centering
    \includegraphics[width=0.99\linewidth]{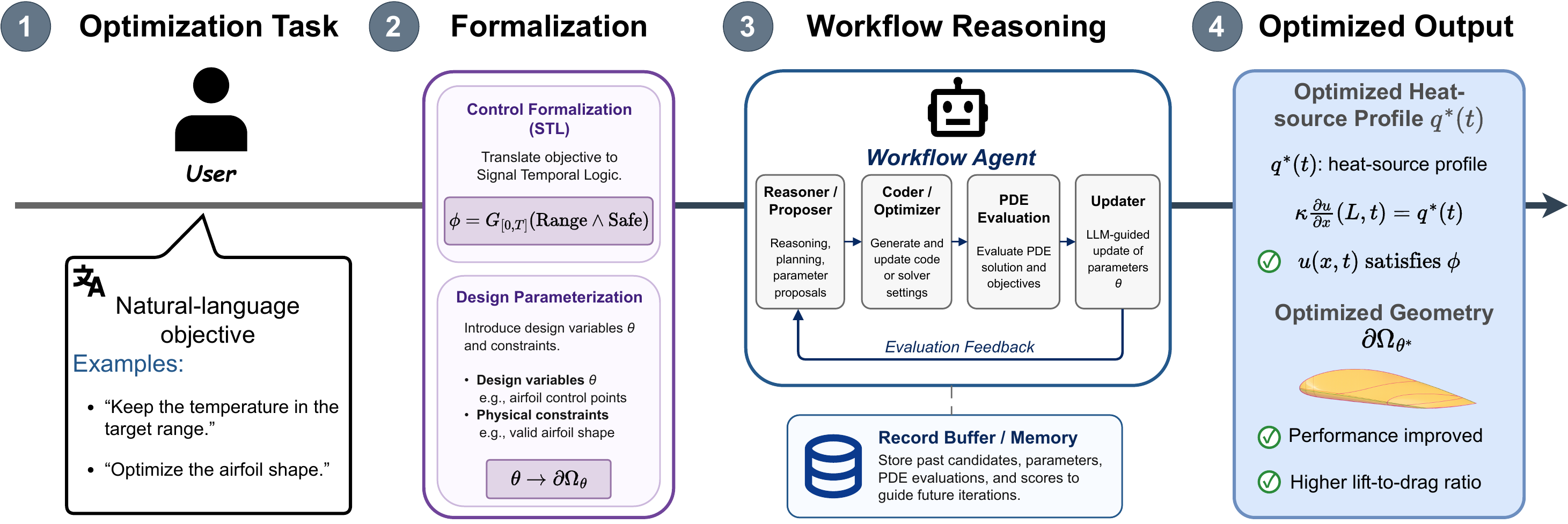}
    \caption{\textbf{LLMs for PDE optimization-stage workflows.}
    A natural-language task is first translated into an explicit objective, constraint set, and decision-space specification. A workflow agent then coordinates computational tools and an iterative search loop in which candidate parameters, control programs, geometries, or model structures are proposed, evaluated through PDE solvers or surrogate models, compared with previous candidates, and revised using physical and task-level feedback. The resulting output can be an optimized control input, such as \(q^*(t)\), or an optimized design, such as \(\partial\Omega_{\lambda^*}\).}
    \label{fig:optimization_stage}
\end{figure}

The Optimization stage concerns optimizing physical parameters, control inputs, geometries, material distributions, or model structures under PDE constraints. Using the notation of Eq.~\eqref{eq:general_pde}, physical design and control can be formulated as
\begin{equation}
\min_{\lambda\in\Lambda}; \mathcal{L}\bigl(u(\lambda),\lambda\bigr)
\quad
\text{subject to}
\quad
\mathcal{F}\left(x,t;u,\partial_tu,\nabla u,\ldots;\lambda\right)=0,
\qquad
\mathcal{C}(u,\lambda)\leq 0,
\label{eq:pde_optimization}
\end{equation}
where \(\lambda\) denotes the decision variables, \(\Lambda\) is the admissible decision space, \(u=u(\lambda)\) is the corresponding PDE state under the prescribed initial and boundary conditions, \(\mathcal{L}\) is the task-level objective, and \(\mathcal{C}\) represents additional physical, operational, or design constraints.

Candidate designs and controls are evaluated through PDE simulations or surrogate predictions of system performance, with the resulting feedback guiding the generation of subsequent candidates. Within this iterative loop, LLMs translate design requirements into computational specifications, coordinate numerical tools, and use evaluation outcomes to refine the search process (Fig.~\ref{fig:optimization_stage}). OptMetaOpenFOAM~\cite{chen2026optmetaopenfoam} organizes this workflow for CFD parameter optimization: its LLM agents identify adjustable parameters and sampling ranges, coordinate simulations and post-processing, and invoke external sensitivity-analysis and optimization tools. Its hydrogen-combustion example seeks the lowest fuel--air equivalence ratio that achieves near-maximal flame propagation distance at a specified time. 

For constraint encoding, engineering requirements are often difficult to express in a form directly usable by numerical optimizers, creating a need to translate them into explicit mathematical conditions and computational inputs. LLMs can facilitate this translation by converting heterogeneous requirements into solver-compatible specifications. PDE-Controller~\cite{soroco2025pdecontroller} addresses temperature and displacement control in systems governed by heat and wave equations. An LLM converts natural-language requirements into signal temporal logic (STL), expressing conditions such as keeping a region's temperature within a prescribed range over a specified interval. A code-generation LLM combines these specifications with discretized PDE dynamics in executable programs, and an external optimizer computes the required time-varying heat inputs or forces. TO-Master~\cite{lin2026tomaster} applies this translation to topology optimization, where material is distributed within a design domain to improve structural stiffness or thermal performance. Its LLM interprets instructions and optional geometry, mesh, or image inputs to specify loads, boundary conditions, material properties, objectives, and volume or stress constraints as arguments for numerical tools. After user confirmation of the mesh and boundary conditions, finite-element and sensitivity-based optimization tools compute the material distribution.

For search guidance, evaluating each candidate can require an expensive PDE simulation, making it important to prioritize designs that are most likely to improve performance. LLMs can support this process by using previous evaluation results to inform the selection and generation of subsequent candidates. In continuous design spaces, LLM-PSO~\cite{zhang2025using} places the LLM inside an iterative search loop for aerodynamic shape optimization of two-dimensional airfoils and three-dimensional axisymmetric bodies. Previously evaluated design--performance pairs are provided to the LLM, which uses them to steer subsequent sampling toward promising regions of the parameter space. TurboAgent~\cite{du2026turboagent} follows a related strategy for compressor blade design, but lets the LLM update both the mean and variance of a sampling distribution from previous parameter--performance pairs, objectives, and constraints, while a surrogate model predicts candidate performance. Under a prescribed mass-flow constraint, the reported surrogate-assisted optimization increased isentropic efficiency from 0.869 to 0.883 and total pressure ratio from 1.589 to 1.637. COSMO-Agent~\cite{deng2026cosmoagent} further learns how such revisions should be made: reinforcement learning trains the LLM to update geometric parameters and material choices for structural components using feedback from CAD generation, finite-element analysis, and cost estimation, with rewards encouraging both constraint satisfaction and efficient tool use.

The same feedback-guided search can operate on executable control policies rather than explicit design parameters. Sun et al.~\cite{sun2026selfevolvingscientificagentdiscovers} study a two-joint, fish-like swimmer that must reach spatial targets. The LLM iteratively revises controller code mapping the swimmer's state and target location to joint angular accelerations, using trajectory summaries and flow visualizations from fluid--structure simulations to diagnose unsuccessful behaviors. During the reported evolution process, the controller improved from reaching one of four optimization targets to reaching all four, and the resulting feedback law subsequently reached all eight static targets reserved for evaluation without further modification. Here, simulation feedback is therefore used to search directly over executable control logic rather than a predefined vector of physical design variables.

Across these examples, the role of the LLM expands from selecting values within a prescribed design space to revising executable policies, physical models, and numerical methods. The underlying pattern nevertheless remains the same: numerical or surrogate evaluation provides feedback that is used to shape the next proposal, allowing the search process to operate over increasingly structured and heterogeneous candidate spaces.

\begin{table*}[t!]
\centering
\footnotesize
\setlength{\tabcolsep}{5pt}
\renewcommand{\arraystretch}{1.12}
\rowcolors{2}{gray!5}{white}

\caption{
Representative benchmarks and evaluation resources for LLM-assisted PDE and scientific-computing workflows.
}
\label{tab:workflow_benchmarks}

\begin{tabularx}{\textwidth}{
>{\raggedright\arraybackslash}p{0.23\textwidth}
>{\raggedright\arraybackslash}p{0.20\textwidth}
>{\raggedright\arraybackslash}X
}

\toprule
\textbf{Benchmark / resource} &
\textbf{Workflow stage} &
\textbf{Evaluation focus} \\
\midrule

LLM-SRBench~\cite{shojaee2025llmsrbench}
&
Discovery: equation discovery
&
239 problems across four scientific domains; evaluates symbolic recovery, numerical fit, and generalization beyond memorized equations.
\\

DiscoverPhysics~\cite{wiemann2026discoverphysics}
&
Discovery: interactive physical-law discovery
&
22 simulated worlds; agents design particle experiments and infer physical laws, evaluated through held-out trajectory error and LLM-judged explanation scores.
\\

Mamo / MM-Bench~\cite{huang2025mamo,liu2025mmagent}
&
Discovery: problem formulation
&
Evaluates natural-language-to-mathematical-model translation, including formulation correctness, computation, and solver-verified results.
\\

CFDLLMBench / FEABench~\cite{somasekharan2025cfdllmbenchbenchmarksuiteevaluating,brenner2024feabench}
&
Solving: numerical software
&
Evaluates CFD and finite-element reasoning, workflow executability, numerical correctness, and complete simulation success.
\\

PDEAgent-Bench / FEM-Bench~\cite{hang2026pdeagentbench,mohammadzadeh2025fembench}
&
Solving: solver code
&
Evaluates solver-code executability, numerical accuracy, efficiency, unit-test generation, and repeated-run reliability.
\\

PINNacle~\cite{PINNacle}
&
Solving: learned solvers
&
Evaluates prediction accuracy, convergence, robustness, and physics-constraint satisfaction across PDE problems.
\\

ShapeBench~\cite{fazliani2026shapebenchscalablebenchmarkdiagnostic}
&
Optimization
&
Evaluates aerodynamic shape optimization using high-fidelity CFD/FEA, including objective improvement, convergence, and feasibility.
\\

ScienceAgentBench~\cite{chen2025scienceagentbench}
&
Cross-stage workflows
&
Evaluates end-to-end scientific task execution, software-environment reconstruction, computational cost, and reproducibility.
\\

SciCode / SciCode-Verified~\cite{tian2024scicode,hu2026scicodeverified}
&
Scientific coding
&
Evaluates research-level scientific code generation and audits benchmark specifications, reference solutions, and graders.
\\

\bottomrule
\end{tabularx}
\end{table*}

\section{Benchmarking and Evaluation across the PDE Workflow}
\label{sec:benchmarking}

Evaluating LLMs in PDE research is challenging because the correctness of a workflow cannot always be determined from its final output alone. Different stages require different forms of verification. In equation discovery, a candidate equation may reproduce observed data without recovering the underlying mathematical structure. In numerical solving, generated code may execute successfully while producing inaccurate solutions, making executability alone insufficient for evaluating solver generation~\cite{hang2026pdeagentbench}. In optimization, improvements in a surrogate objective may not persist under high-fidelity simulation. Reliable evaluation therefore requires benchmarks that examine not only task completion, but also the mathematical, numerical, and physical validity of the resulting workflow. Constructing such benchmarks is itself demanding, as representative PDE problems, verified reference solutions, and reliable evaluation procedures require substantial domain expertise. In this section, we review existing benchmarks and evaluation frameworks for the Discovery, Solving, and Optimization stages (Table~\ref{tab:workflow_benchmarks}) and examine the capabilities they measure and the limitations of current evaluation practices.

\subsection{Evaluation in the Discovery Stage}
\label{sec:benchmark_discovery}

In the Discovery stage, evaluation first asks whether a system can recover the correct scientific relation from fixed observations. LLM-SRBench~\cite{shojaee2025llmsrbench} provides 239 equation-discovery problems comprising transformed physical equations and synthetic equations from chemistry, biology, physics, and materials science. Given numerical observations and descriptions of the variables, a system must generate an equation explaining their relationship. Evaluation considers both numerical prediction error and recovery of the target symbolic structure, separating the ability to fit observations from the ability to recover the underlying mathematical form. Symbolic equivalence is assessed by an LLM judge after abstracting fitted constants, with expert checks used to validate the resulting judgments. This distinction is important because a numerically accurate expression can still represent an incorrect scientific law.

A broader level of discovery evaluation asks whether an agent can actively determine what evidence should be collected before proposing a law. DiscoverPhysics~\cite{wiemann2026discoverphysics} evaluates this capability across 22 simulated worlds governed by unfamiliar physical rules. Rather than receiving a fixed dataset, agents design particle experiments, observe the resulting trajectories, revise their hypotheses, and finally submit both a textual explanation and an executable implementation of the inferred law. Prediction on held-out trajectories is combined with rubric-based evaluation of the scientific explanation. Although DiscoverPhysics is not PDE-specific, it captures an important capability that fixed-data equation-discovery benchmarks do not: the ability to choose informative observations and use them to refine a scientific hypothesis.

Problem formulation introduces a different evaluation problem: whether an informal description can be converted into an executable mathematical model. Mamo~\cite{huang2025mamo} contains 1,209 problems involving ordinary differential equations, linear programming, and mixed-integer linear programming. Generated formulations are executed by numerical or optimization solvers, allowing the resulting answers to be compared with reference solutions. MM-Bench~\cite{liu2025mmagent} extends this setting to 111 open-ended mathematical-modelling problems drawn from real-world modelling competitions. Systems must analyse the problem, construct a mathematical model, perform the required computation, and explain the resulting conclusions, with rubric-based evaluation supplemented by expert assessment. These resources provide useful evidence for language-to-model formulation, but neither is dedicated to PDE formulation. Dedicated benchmarks that test whether LLMs can translate scientific descriptions into complete governing PDEs, initial and boundary conditions, constitutive assumptions, and modelling constraints therefore remain scarce.

Overall, current Discovery-stage benchmarks evaluate three increasingly demanding capabilities: recovering a mathematical relation from fixed observations, actively acquiring evidence through experimentation, and constructing a model from an informal problem description. These capabilities are currently evaluated in largely separate settings, and a benchmark that integrates them within realistic PDE discovery remains an important missing component.

\subsection{Evaluation in the Solving Stage}
\label{sec:benchmark_solving}

Solving-stage evaluation needs to establish not only whether generated code can run, but also whether it produces accurate and convergent numerical solutions at reasonable computational cost. CFDLLMBench~\cite{somasekharan2025cfdllmbenchbenchmarksuiteevaluating} evaluates generated PDE programs on 24 one- and two-dimensional problems by comparing them with analytical solutions or expert implementations and examining convergence under mesh or time-step refinement. Its OpenFOAM track contains 126 additional cases and compares computed physical quantities with reference simulations. FEABench~\cite{brenner2024feabench} evaluates COMSOL workflow generation for finite-element and multiphysics problems. Among its tasks, 15 manually verified Gold problems provide quantitative reference results, while 200 tutorial-derived cases broaden task coverage. These results make clear that successful execution alone provides limited evidence of scientific correctness.

PDEAgent-Bench~\cite{hang2026pdeagentbench} focuses more directly on solver generation, with 645 instances from 11 PDE families evaluated by execution success, solution error, and computational efficiency. FEM-Bench~\cite{mohammadzadeh2025fembench} examines lower-level numerical programming through 33 computational-mechanics tasks for function and unit-test generation. For the latter, generated tests must accept a correct reference implementation while detecting predefined faulty implementations, thereby testing whether the model can verify numerical code as well as produce it.

For LLM-assisted learned solvers, numerical benchmarks can instead serve as testbeds for model and training choices proposed by the LLM. PINNacle~\cite{PINNacle} contains more than 20 challenging PDE problems for physics-informed neural networks, including complex geometries, multiscale behaviour, and nonlinear dynamics. Although it is not an LLM benchmark, it can be used to compare proposed architectures, loss functions, sampling strategies, and training procedures.

Current benchmarks therefore cover execution, numerical accuracy, convergence, efficiency, and code verification, but these aspects are still largely assessed separately. An end-to-end benchmark spanning method selection, implementation, diagnosis, correction, and final numerical verification remains lacking.

\subsection{Evaluation in the Optimization Stage}
\label{sec:benchmark_optimization}

In the Optimization stage, evaluation asks whether an agent can identify better feasible candidates under a controlled search budget. ShapeBench~\cite{fazliani2026shapebenchscalablebenchmarkdiagnostic} provides 103 aerodynamic shape-optimization tasks across eight categories with standardized design spaces, objectives, and evaluators. Classical and LLM-based optimizers can therefore be compared under matched numbers of candidate evaluations. Search is primarily performed using surrogate models, while supported tasks can additionally be checked using high-fidelity CFD or FEA. This distinction is particularly important because optimization can exploit inaccuracies in a surrogate model: an unusual geometry may receive an artificially favorable predicted objective while performing substantially worse under higher-fidelity simulation. Evaluation of PDE-constrained optimization should therefore report objective improvement and constraint satisfaction under a common evaluator, together with computational budget and final validation using a shared high-fidelity model where available. Furthermore, existing aerodynamic geometry datasets can also provide complementary test beds for evaluating LLM-based optimizers, including AirFoil~\cite{ramos2023airfoil9k}, DrivAerNet++~\cite{elrefaie2024drivaernet}, and DrivAerML~\cite{ashton2024drivaerml}.

Optimization-stage evaluation is consequently different from simply measuring whether an objective decreases. A credible improvement must remain valid under the governing physical constraints and survive evaluation with a numerical model whose fidelity is sufficient to distinguish genuine design improvement from exploitation of approximation error.

\subsection{Cross-workflow Evaluation and Benchmark Reliability}
\label{sec:benchmark_cross_workflow}

Beyond benchmarks designed specifically for individual PDE stages, broader scientific-agent benchmarks can evaluate whether PDE-related computational reasoning remains reliable when embedded in longer scientific workflows. ScienceAgentBench~\cite{chen2025scienceagentbench} contains 102 tasks derived from 44 peer-reviewed publications across four scientific disciplines. Given task instructions and associated data, agents generate executable programs for scientific analysis and visualization. Its relevance to PDE research lies in tasks that require agents to invoke existing simulation software, process numerical outputs, and connect computational results to the requested scientific analysis. The benchmark therefore evaluates capabilities that also arise in PDE workflows, including tool use, data handling, program execution, and result extraction, although it is not itself a PDE-specific benchmark.

SciCode~\cite{tian2024scicode} provides a more direct test of PDE-related scientific programming. It contains research-level coding problems across 16 natural-science subfields, including tasks involving equations such as Burgers, heat, Maxwell, and Schrödinger equations, where models must translate scientific specifications into numerically correct implementations. SciCode-Verified~\cite{hu2026scicodeverified} retains this scientific-programming focus while auditing and correcting the benchmark itself. The authors identified 192 defects capable of rejecting otherwise correct solutions across 58 of the 64 problems retained in the corrected release, including errors in reference outputs, numerical tolerances, task specifications, and grading procedures. Because PDE code is often judged through numerical tests, this audit is particularly relevant to PDE evaluation: an incorrect reference solution, unstable tolerance, or flawed test condition can confound genuine differences in solver correctness and numerical accuracy.

Across the workflow, benchmark development is therefore progressing along two dimensions. The first is increasing scientific depth, from task completion and code execution toward mathematical correctness, numerical convergence, physical validity, optimization fidelity, and computational reproducibility. The second is increasing workflow scope, from isolated equation or coding tasks toward multi-step interactions involving model construction, numerical software, diagnosis, and result verification. For PDE-oriented evaluation, comparisons should consequently report not only final success rates, but also the information and tools available to the agent, the numerical solver and software versions, repeated-attempt budget, computational cost, expert intervention, and the procedures used to validate reference solutions and graders.

\section{Challenges and Outlook}

The studies reviewed above show that LLMs are increasingly involved across the PDE research workflow, from equation discovery and numerical solving to optimization. Most existing systems still operate within established problem settings and predefined computational procedures, with the main scientific choices and evaluation criteria largely specified in advance. Further progress therefore depends on extending the role of LLMs from executing predefined tasks to contributing more directly to the reasoning, learning, and judgement required in scientific research. Three directions are particularly important: mathematical and physical reasoning for identifying the essential structure of PDE problems and selecting appropriate solution strategies; domain-specific post-training for developing transferable expertise across equations, numerical methods, and computational environments; and scientific judgement for assessing computational results when explicit reference answers are unavailable.

First, a central challenge is to develop mathematical and physical reasoning that adapts to the structure of each PDE problem. This requires LLMs to identify appropriate assumptions, select relevant principles, and construct derivations that remain consistent with the governing physics and mathematical constraints. Developing this capability opens opportunities for LLMs to contribute to governing equations, numerical schemes, and optimization strategies across PDE workflows. At the discovery stage, much of the work reviewed here uses LLMs to construct candidate equations or guide efficient searches~\cite{yang2026thinklikescientistphysicsguided,grayeli2024lasr}. A promising next step is to combine these data-driven approaches with problem-specific mathematical reasoning, enabling LLMs to undertake aspects of the work performed by mathematicians and physicists: formulating physical hypotheses, identifying appropriate assumptions, and deriving governing relations from conservation laws, symmetries, and constitutive descriptions. Observational data can then help identify unresolved terms, distinguish competing hypotheses, and refine the resulting models. At the solving stage, reasoning can extend the role of LLMs from invoking numerical tools and implementing established methods to developing new numerical schemes tailored to a problem's mathematical structure and physical behaviour. Analysing conservation properties, dominant balances, and interacting scales can inform the construction of discretizations and time-integration schemes, with their consistency, stability, and computational efficiency assessed through mathematical analysis and numerical experiments. At the optimization stage, similar reasoning can connect design or control choices to the physical mechanisms governing performance, motivating new design parameterizations, structural configurations, or feedback laws. Simulation results and sensitivity information can then be used to test these mechanistic hypotheses and refine the proposed strategies. Across these stages, an important research goal is to enable LLMs to formulate and revise their own analytical approach, including recognizing when an assumption no longer applies or when a different mathematical representation is needed.

Second, domain-specific post-training can address the challenge of learning transferable PDE knowledge from heterogeneous scientific resources. Many existing approaches use general-purpose pretrained models through prompting and tool integration, creating an opportunity to complement these workflows with models explicitly trained on PDE knowledge and reasoning. The domain provides rich resources, including textbooks, mathematical derivations, numerical-analysis literature, solver source code, and software documentation. Aligning these resources requires preserving the physical assumptions, initial and boundary conditions, and numerical settings that connect a governing equation to its derivation and implementation. Organizing such relationships into high-quality training examples provides a foundation for domain specialization. Existing studies illustrate this direction: Dong et al. fine-tune a language model on the NL2FOAM dataset, which pairs natural-language descriptions with OpenFOAM configurations, while ALL-FEM uses verified FEniCS scripts to specialize language models for finite-element workflows~\cite{Dong2025,DEOTALE2026118985}. A complementary opportunity lies in multimodal post-training~\cite{cao2026learningdomainknowledgemultimodal}. Vision--language models can be fine-tuned on snapshots and temporal sequences of PDE solution fields, paired with governing equations, initial and boundary conditions, simulation parameters, and physical interpretations. These training examples must retain the spatial scales, temporal information, and physical quantities needed to interpret the visual patterns. This alignment can support the interpretation of simulation results and diagnosis of numerical artefacts. A key research question is how to organize training across textual, symbolic, visual, and executable representations so that models learn relationships that transfer across equation families, numerical methods, and software environments. Evaluation on unfamiliar combinations of these elements is essential for establishing the scope of the resulting PDE expertise.

Third, open-ended PDE problems pose the challenge of determining how a new hypothesis, numerical method, or design should be tested when reference solutions or established evaluation procedures are unavailable. The choice of experiment becomes part of the scientific problem: different tests probe different aspects of a claim, and discrepancies may arise from physical assumptions, numerical approximations, or measurement uncertainty. Active experimental design provides a way to address this challenge by selecting simulations and measurements that distinguish competing explanations and clarify the conditions under which a conclusion holds. Existing systems illustrate this potential: Coscientist integrates experimental planning and execution in chemistry, while The AI Scientist generates research ideas, conducts computational experiments, and analyses their outcomes~\cite{Boiko2023Coscientist,Lu2026AIScientist}. For PDE research, a promising direction is to enable LLMs to design numerical and physical experiments around specific scientific questions. In equation discovery, this involves selecting measurement locations, forcing conditions, or parameter regimes where competing hypotheses predict different behaviour. In numerical solving, it involves devising test problems, conducting refinement studies, and comparing independent implementations to examine the accuracy, stability, and conservation properties of a proposed method. In optimization, it involves testing designs or control strategies under new operating conditions and higher-fidelity simulations to assess their robustness and practical value. Each experiment can inform the next: unexpected observations can motivate a revised physical hypothesis, additional measurements, or a more detailed numerical analysis. Developing this capability requires LLMs to interpret the outcomes of different tests, identify unresolved questions, and prioritize follow-up experiments according to their scientific purpose and computational cost. Integrating experimental design with scientific validation can help LLMs investigate new phenomena, establish the applicability of new methods, and refine scientific conclusions across PDE workflows.

\bibliographystyle{unsrt}
\bibliography{references}







\end{document}